\documentclass[10pt,twocolumn,letterpaper]{article}

\usepackage{iccv}
\usepackage{times}
\usepackage{epsfig}
\usepackage{graphicx}
\usepackage{amsmath}
\usepackage{amssymb}
\usepackage{breqn}
\usepackage[breaklinks=true,bookmarks=false]{hyperref}

\iccvfinalcopy % *** Uncomment this line for the final submission

\def\iccvPaperID{****} % *** Enter the ICCV Paper ID here

\ificcvfinal\fi

\begin{document}

%%%%%%%%% TITLE
\title{Learning Bag-of-Features Pooling for Deep Convolutional Neural Networks}

\author{Nikolaos Passalis and  Anastasios Tefas\\
Aristotle University of Thessaloniki\\
Thessaloniki, Greece\\
{\tt\small passalis@csd.auth.gr, tefas@aiia.csd.auth.gr}
%\and
%Anastasios Tefas \\
%Aristotle University of Thessaloniki\\
%Thessaloniki, Greece\\
%{\tt\small tefas@aiia.csd.auth.gr}
}

\maketitle
\thispagestyle{empty}

%%%%%%%%% ABSTRACT
\begin{abstract}
	Convolutional Neural Networks (CNNs) are well established models capable of achieving state-of-the-art classification accuracy for various computer vision tasks. However, they are becoming increasingly larger, using millions of parameters, while they are restricted to handling images of fixed size. In this paper, a quantization-based approach, inspired from the well-known Bag-of-Features model, is proposed to overcome these limitations. The proposed approach, called Convolutional BoF (CBoF), uses RBF neurons to quantize the information extracted from the convolutional layers and it is able to natively classify images of various sizes as well as to significantly reduce the number of parameters in the network. In contrast to other global pooling operators and CNN compression techniques the proposed method utilizes a trainable pooling layer that it is end-to-end differentiable,  allowing the network to be trained using regular back-propagation and to achieve greater distribution shift invariance than competitive methods. The ability of the proposed method to reduce the parameters of the network and increase the classification accuracy over other state-of-the-art techniques is demonstrated using three image datasets.
\end{abstract}

\section{Introduction}
Deep Convolutional Neural Networks (CNNs) are powerful well-known models used for various computer vision tasks, ranging from image classification \cite{russakovsky2015imagenet}, to visual question answering \cite{ma2015learning}. Their great success in difficult large-scale recognition problems, such as classifying the images of the ImageNet dataset \cite{russakovsky2015imagenet}, established  them as the model of choice for the aforementioned tasks, displacing other models that were previously used, such as the Bag-of-Features (BoF) model, also known as the Bag-of-Visual-Words (BoVW) model \cite{lazebnik2006beyond}.

CNNs are composed of a feature extraction layer followed by a fully connected layer that acts as a classifier. The feature extraction layer is further composed of a series of convolutional layers and pooling layers. Many different combinations of convolution and pooling operators have been proposed \cite{he2015deep, he2014spatial, lecun1998gradient, simonyan2014very}. The feature maps extracted from the last convolutional/pooling layer are flatten into a vector which is subsequently fed to the fully connected layer. Regardless the details of the used architecture, all the CNNs share the same fundamental principle: the parameters of the feature extraction layer are not fixed, but they are learned using back-propagation, just as any other parameter of the network.

Despite their great success, most of the proposed CNN formulations are unable to handle arbitrary sized images, as they operate on a fixed input size, restricting the CNN into accepting images of a specific size and leading to a constant cost for feed-forwarding the network (the number of floating point operations (FLOPs) needed to feed-forward the network depends on the size of the input image). This is because the dimensionality of the (flattened) output of the feature extraction layer depends on the size of the input image. Therefore, an already trained CNN cannot be directly deployed in a lower computational resource scenario, \ie, an embedded real-time system on a drone or a mobile device, or adapt to the available computational resources, without completely replacing the fully connected layer and retraining it. Furthermore, flattening the extracted feature maps leads to tremendously large fully connected layers. For example, for the VGG-16 network  \cite{simonyan2014very}, 90\% of the network's parameters are needed for just three fully connected layers, while only 10\% of the parameters are actually spent on  its 13 convolutional layers used for the feature extraction. The reason for this is the large number of the extracted feature maps (512 feature maps of size 7x7).

There were some attempts to overcome the aforementioned limitations by using global pooling operators that ensure that the output of the feature extraction layer has fixed length \cite{he2014spatial, oquab2014learning}. Even though these global pooling operators can be used to provide scale invariance and allow the network to operate with arbitrary sized images, it was experimentally established (Section \ref{section:experiments}) that the accuracy of the network is reduced when the input image size is altered. This is because the scale invariance is not provided by learning a scale-invariant pooling layer, but by learning scale-invariant filters that are then pooled together. However, as it is demonstrated in this work, learning both the pooling layer and the convolutional layers can provide significantly better scale-invariance, while reducing the size of the resulting network. Note that compression techniques, such as \cite{gong2014compressing, han2015deep, wu2015quantized}, can be also used to reduce the size of CNNs, however they are not capable of dealing with arbitrary sized images.

In this work, a BoF-inspired layer, that acts as a trainable quantization-based pooling layer, is used between the feature extraction layer and the fully connected layer to resolve these problems. It worths mentioning that the BoF model was originally designed to tackle similar problems, i.e., to deal with a variable number of feature vectors and provide scale and position invariance, that arose when handcrafted feature extractors were used. To understand this, consider the pipeline of the BoF model: First, a number of feature vectors are extracted from an image using a handcrafted feature extractor, such as SIFT \cite{lowe1999object}, HoG \cite{dalal2005histograms}, or LBP \cite{ojala2002multiresolution}. The number of feature vectors might vary according to the type of feature extraction and the size of each image. Then, these feature vectors are quantized into a predefined number of bins, called codewords. Finally, a constant length histogram representation is extracted for each image by counting the number of feature vectors that were quantized into each bin.

\begin{figure*}
	\begin{center}
		\includegraphics[width=0.99\linewidth]{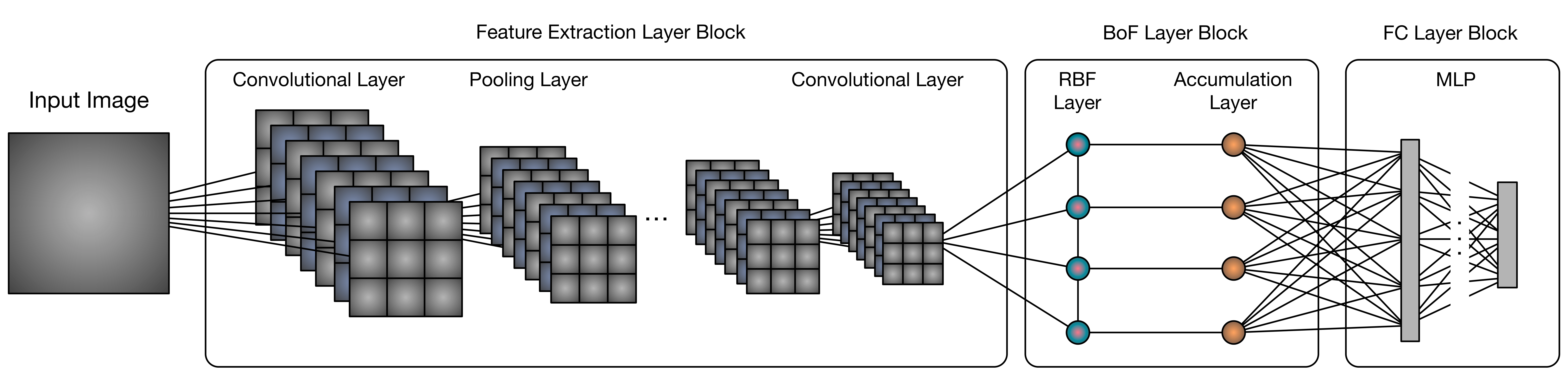}
	\end{center}
	\caption{Convolutional BoF Model}
	\label{fig:cbof}
\end{figure*}

The feature maps extracted from the last convolutional layer of a CNN can be converted to feature vectors as follows. Each convolutional filter contributes to a specific position of the feature vectors, i.e., the dimensionality of the extracted feature vectors equals to the number of the used convolutional filters, while the size of the feature maps defines the number of the extracted feature vectors. Instead of directly feeding the extracted vectors into the fully connected layer, a BoF-based pooling layer is used between the last convolutional layer and the fully connected layer. This can reduce the size of the fully connected layer and allows to operate the CNN using images of any size. However, combining the BoF model with CNNs is not a straightforward task. The representation learned by the convolutional layers of a CNN is constantly altering during the training prohibiting standard BoF quantization techniques, such as k-means quantization, to be used effectively. Therefore, the BoF model must be appropriately modified in order to efficiently handle trainable convolutional feature extractors.

The contributions of this work are briefly summarized bellow. First, a neural generalization of the BoF model, composed of RBF neurons, is proposed as a generic pooling layer that can fit between the feature extraction layer and the fully connected layer of a CNN. Note that the proposed pooling layer can be also used at various depths of the network to further increase the scale-invariace and bring more fine-grained information to the fully connected layer. To the best of our knowledge, this is the first time that the BoF model is combined with convolutional neural networks into one unified architecture, called Convolutional BoF (CBoF), that is end-to-end differentiable and allows from training the resulting network using regular back-propagation. It is shown, using extensive experiments on three datasets, that CBoF provides much better invariance to distribution shifts, caused by feeding images of different sizes to the network than competitive global pooling strategies, such as Spatial Pyramid Pooling~\cite{he2014spatial}. Also it can greatly reduce the number of parameter of the network over the competitive methods. This allows the proposed technique to be easily used with embedded devices with limited processing power and memory, e.g., GPU-based processing units for drones and robots. Finally, a spatial segmentation scheme is also proposed and combined with the CBoF to retain the spatial information that is carried by the extracted feature vectors. A reference implementation of the proposed method is available at \url{https://github.com/passalis/cbof}.

The rest of the paper is structured as follows. In Section~\ref{section:rel-work} the related work is introduced and compared to the proposed CBoF model. The proposed method is presented in detail in Section~\ref{section:proposed-method}. Next, the CBoF is evaluated and compared to other state-of-the-art techniques using three different datasets (Section~\ref{section:experiments}), including a large-scale face image dataset. Finally, conclusions are drawn in Section~\ref{section:conclusions}.

\section{Related Work}
\label{section:rel-work}

The problem of dealing with large CNNs is well recognized in the literature, with several recent works trying to reduce the model size \cite{gong2014compressing, han2015deep, he2014spatial, wu2015quantized}. Most of these works use compression and pruning techniques to reduce the size of  CNN models \cite{chen2015compressing, gong2014compressing, han2015deep, wu2015quantized}. These approaches focus on compressing an already trained CNN, instead of training a CNN with less parameters in the first place. Also they are unable to natively handle images of various sizes, leading to a constant cost for feed-forwarding the convolutional layers (since the image must be resized to a predefined size). 
Note that vector quantization is used both by some of these works \cite{gong2014compressing, wu2015quantized}, and the proposed CBoF technique. However, the CBoF method uses a differentiable quantization scheme that allows training both the quantizer and the rest of the network simultaneously, instead of using fixed quantization just to reduce the size of the model. This allows for directly training CNN models with less parameters (instead of compressing them after training) as well as natively handling differently sized images.  Note that the proposed CBoF model can be readily combined with any CNN  compression technique to further decrease the size of the model.

The problem of dealing with arbitrary sized images is addressed in the literature using global pooling operators \cite{he2014spatial, oquab2014learning, malinowski2013learnable}. Since the naive global max pooling \cite{oquab2014learning}, leads to the loss of valuable spatial information, spatial pooling techniques \cite{malinowski2013learnable, he2014spatial}, were proposed to overcome this issue. Both the global pooling techniques and the proposed CBoF technique reduce the size of network and they can handle images of different sizes. However, as we show in Section~\ref{section:experiments}, the proposed method greatly outperforms all the other global pooling methods, achieving both higher recognition accuracy and using overall less parameters. This can be attributed to the ability of the proposed method to \textit{learn} how to perform quantization-based pooling, that allows for efficiently compressing the output of the feature extraction layer and providing much better invariance to distribution shifts caused by feeding differently sized images to the network.

Finally, the proposed method is related to supervised dictionary learning approaches for the BoF representation. A very rich literature exists for BoF supervised dictionary learning \cite{lazebnik2009supervised, lian2010max, lobel2014joint, passalis2017neural}. However, all these methods are designed to work with handcrafted feature extractors instead of trainable convolutional layers. To the best of our knowledge, this is the first work that combines the BoF model with convolutional layers into one unified architecture that allows for training the resulting network from scratch by back-propagating the error from the fully connected layer to the convolutional feature extractor.

\section{Proposed Method}
\label{section:proposed-method}

The proposed Convolutional BoF model is composed of three layer blocks: a) a feature extraction layer block (composed of many convolutional and pooling layers), b) a BoF pooling layer block and c) a fully connected layer block. The structure of the CBoF model is illustrated in Figure \ref{fig:cbof}. Each of the layers used in the CBoF is described in detail in the next subsections. Next, a learning algorithm for the CBoF model is derived and its computational complexity is discussed and compared to other methods.

\subsection{Feature Extraction Layer Block}
Let $\mathcal{X}$ be a set of $N$ images to be classified using the CBoF model. The $i$-th  image is fed to the feature extraction layer, which is composed of a sequence of convolutional layers and subsampling (pooling) layers. Any CNN architecture can be used as feature extractor, such as the LeNet~\cite{lecun1998gradient}, the VGG~\cite{simonyan2014very}, or the ResNet~\cite{he2015deep}, after removing its fully connected layers.

The last convolutional layer, denoted by $L$, is used to extract feature vectors that are subsequently fed to the BoF layer. The $j$-th feature vector of the $i$-th image is denoted by $\mathbf{x}_{ij} \in \mathbb{R}^{D}$, where $D$ is the number of filters used in the last convolutional layer. The number of the extracted feature vectors depends on the size of the feature map and the used filter size, as described in the Introduction. For example, for an image of 28x28 pixels and two convolutional layers with filter size 5x5 a total of $20 \times 20 = 400$ feature vectors can be extracted from the final convolutional layer (assuming that the filters fully overlap with the image). To simplify the presentation of the proposed method, the number of feature vectors extracted from the last convolutional layer for the $i$-th image is denoted by $N_i$.

\subsection{BoF Layer Block}
After the feature extraction the $i$-th image is represented by a set of $N_i$ feature vectors: $\mathbf{x}_{ij} \in \mathbb{R}^D$ ($j = 1...N_i)$.  Instead of simply fusing the extracted feature vectors, like the CNNs \cite{he2015deep, lecun1998gradient, russakovsky2015imagenet, simonyan2014very}, or using a spatial pooling approach, like the SPP \cite{he2014spatial}, the BoF layer compiles a fixed-length histogram for each image by quantizing its feature vectors into a predefined number of histogram bins/codewords. Note that the length of the extracted histogram vector does not depend on the number of available feature vectors, which allows the network to handle images of arbitrary size without any modification.

The BoF model is formulated as a generic neural layer that is composed of two sublayers: an RBF layer that measures the similarity of the input features to the RBF centers and an accumulation layer that builds the histogram of the quantized feature vectors. The proposed layer can be thought as a unified processing layer that feeds the extracted representation to a subsequent classifier. The output of the $k$-th RBF neuron $[\boldsymbol{\phi}(\mathbf{x})]_k$  is defined as:
\begin{equation}
\label{eq:new-similarity-eq}
[\boldsymbol{\phi}(\mathbf{x})]_k = \exp(-||(\mathbf{x}-\mathbf{v}_k) ||_2 / \sigma_k)
\end{equation}
where $\mathbf{x}$ is a feature vector and $\mathbf{v}_k$ is the center of the $k$-th RBF neuron. The RBF neurons behave somewhat like the codewords in the BoF model, \ie, they are used to measure the similarity of the input vectors to a set of predefined vectors and quantize the feature space. Each RBF neuron is also equipped with a scaling factor $\sigma_k$ that adjusts the width of its Gaussian function. That allows for better modeling of the input distribution, since the distribution modeled by each RBF can be independently learned. The number of RBF neurons used is denoted by $N_K$. The size of the extracted representation can be adjusted by using a different number of RBF neurons ($N_K$) in the BoF layer.

To ensure that the output of each RBF neuron is bounded, a normalized RBF architecture is used. This normalization is equivalent to the $l^1$ scaling that is utilized in the BoF model that uses soft-assignments \cite{van2010visual}. Thus, the output of the RBF neurons is re-defined as:
\begin{equation}
[\boldsymbol{\phi}(\mathbf{\mathbf{x}})]_{k} = \frac{\exp(-||(\mathbf{x}-\mathbf{v}_k) ||_2 / \sigma_k)}{\sum_{m=1}^{N_K}\exp(-||(\mathbf{x}-\mathbf{v}_m) ||_2 / \sigma_m)}
\end{equation}

The output of the RBF neurons is accumulated in the next layer, compiling the final representation of each image:
\begin{equation}
\label{eq:input-layer-output}
\mathbf{s}_i = \frac{1}{N_i}\sum_{j=1}^{N_i} \boldsymbol{{\phi}}(\mathbf{x}_{ij}) 
\end{equation}
where $\boldsymbol{{\phi}}(\mathbf{x}, t) = ([{\boldsymbol{\phi}}(\mathbf{x}, t)]_1, ..., [{\boldsymbol{\phi}}(\mathbf{x}, t)]_{N_K})^T  \in \mathbb{R}^{N_K} $ is the output vector of the RBF layer. Note that each $\mathbf{s}_i$ has unit $l^1$ norm and defines a histogram distribution over the RBF neurons that describes the visual content of each image. The vector $\mathbf{s}_i$ can be then used for the subsequent classification or retrieval tasks \cite{passalis2016entropy}. Hard quantization is also supported by the proposed formulation. In this case, only the RBF neuron with the maximum response is activated for each feature vector and the rest of the architecture remains unchanged. However, this modification introduces non-continuities that make the optimization of the input layer intractable and it is not used in this work. 

\begin{figure}
	\begin{center}
		\includegraphics[width=0.7\linewidth]{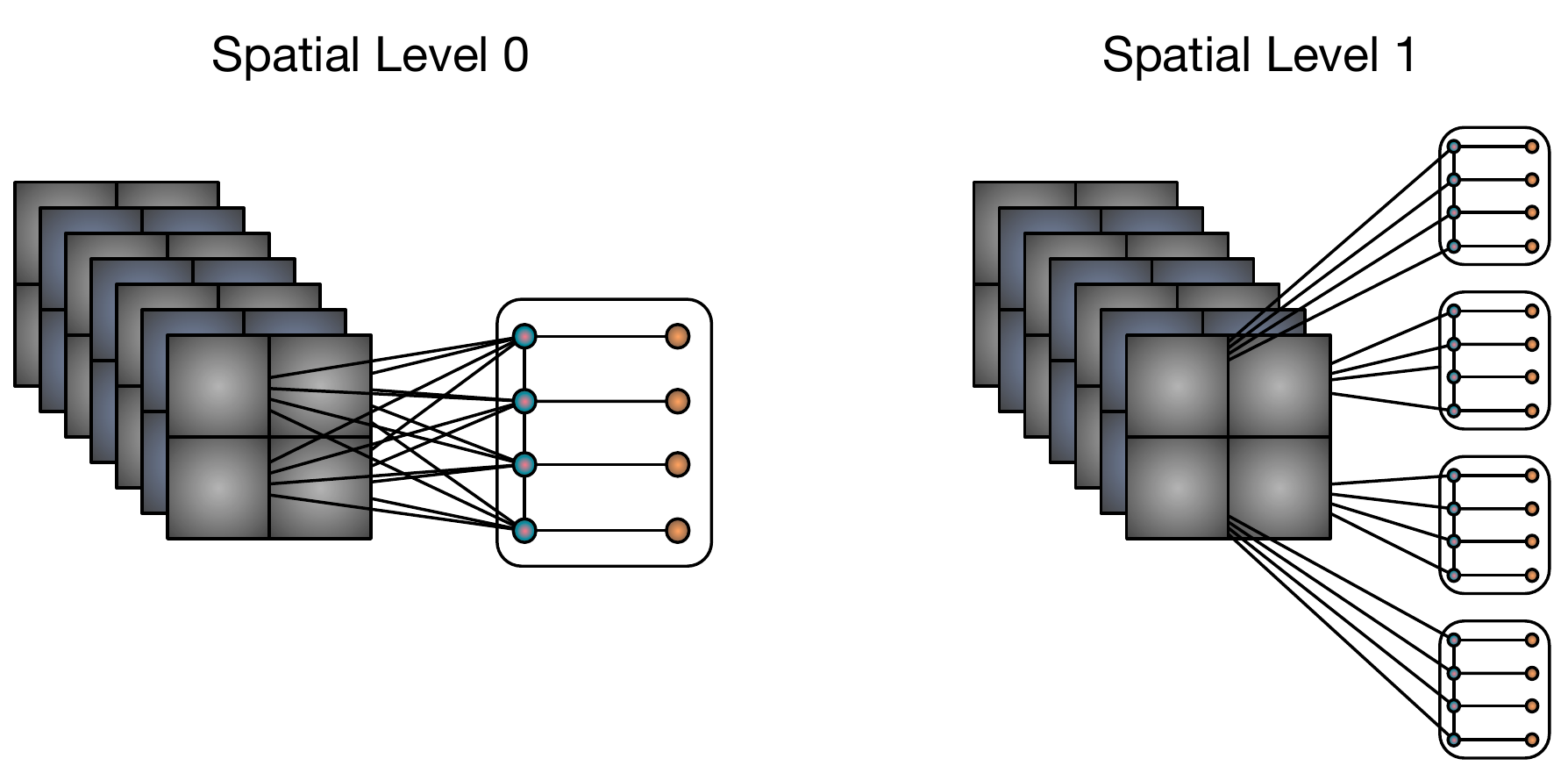}
	\end{center}
	\caption{Using a spatial segmentation scheme to introduce spatial information into the extracted CBoF representation.}
	\label{fig:cbof-spatial}
\end{figure}

Note that the previous process discards most of the spatial information carried by the feature vectors, since all the feature vectors are described by the same histogram regardless the part of the image from which they were extracted. To overcome this, a spatial segmentation scheme, similar to the Spatial Pyramid Matching scheme \cite{lazebnik2006beyond}, is proposed. Before quantizing the feature vectors using the BoF layer, each feature vector is assigned into one spatial region and a separate BoF layer is used to quantize the feature vectors that belong to each region. That way, a different histogram is extracted from each region, allowing to introduce more spatial information into the CBoF model. This process is illustrated in Figure~\ref{fig:cbof-spatial}. The number of used spatial regions is denoted by $N_S$.

\subsection{Fully Connected Layer}

The previous layer receives the feature vectors of an image and compiles its histogram representation. Then, this histogram must be fed to a classifier that decides the class of the image. In this work a multilayer perceptron (MLP) with one hidden layer is used for this purpose, although any other classifier with differentiable loss function can be used.

Only the single-label classification problem is considered: the $i$-th training image  is annotated by a label $t_i$ and there are $N_C$ different labels. This is without loss of generality since the proposed method can be readily applied to multi-label classification or regression problems as well.

Let $\mathbf{W}_H \in \mathbb{R}^{N_H \times (N_KN_S)}$ be the hidden layer weights and $\mathbf{W}_O \in \mathbb{R}^{N_C \times N_H}$ be the output layer weights, where $N_H$ is the number of hidden neurons.  Then, the hidden layer activations for the input histogram  $\mathbf{s}_i$ of the $i$-th image are computed as:
\begin{equation}
\label{eq:hidden-output}
\mathbf{h}_i = \phi^{(elu)}(\mathbf{W}_H \mathbf{s}_i + \mathbf{b}_H) \in \mathbb{R}^{N_H}
\end{equation}
where $\phi^{(elu)}(x)$ is the \textit{elu} activation function \cite{clevert2015fast}: $\phi^{(elu)}(x) = x$ if $x>0$, or $\phi^{(elu)}(x) = \alpha_{elu} (exp(x)-1)$ otherwise. The parameter $\alpha_{elu}$ is typically set to 1. The activation function is applied element-wise and $\mathbf{b}_H \in \mathbb{R}^{N_H}$ is the hidden layer bias vector.

Similarly, the output of the MLP is calculated as:
\begin{equation}
\label{eq:mlp-output}
\mathbf{y}_i = \phi^{(softmax)}(\mathbf{W}_O \mathbf{h}_i + \mathbf{b}_O) \in \mathbb{R}^{N_C}
\end{equation}
where each output neuron corresponds to a label (the one-vs-all strategy is used), $\mathbf{b}_O \in \mathbb{R}^{N_C}$ is the output layer bias vector and $\phi^{(softmax)}$ is the softmax activation function.

In this work, the categorical cross entropy loss is used for training the network:
\begin{equation}
\label{eq:loss}
L = -\sum_{i=1}^N  \sum_{j=1}^{N_C} [\textbf{t}_{i}]_j \log( [\textbf{y}_{i}]_j)
\end{equation}
where $\mathbf{t}_i \in \mathbb{R}^{N_C}$ is the target output vector, which depends on the label ($t_i$) of the input image and it is defined as: $[\textbf{t}_{i}]_j = 1$, if $j=t_i$, or $[\textbf{t}_{i}]_j = 0$, otherwise. For both the input layer and the hidden layer the dropout technique can be used \cite{srivastava2014dropout}. Dropout with rate $p=0.5$ is used, except otherwise stated.

\subsection {Learning with the CBoF}

All the layers of the CBoF network can be trained using back-propagation and gradient descent:
$ \Delta(\mathbf{W}_{MLP}, \mathbf{V}, \boldsymbol{\sigma}, \mathbf{W}_{conv}) = - (\eta_{MLP} \frac{\partial L}{\partial \mathbf{W}_{MLP}},
	\eta_{V} \frac{\partial L}{\partial \mathbf{V}},
	\eta_{\sigma} \frac{\partial L}{\partial \boldsymbol{\sigma}},
	\eta_{conv} \frac{\partial L}{\partial \mathbf{W}_{conv}})$, 
where the notation $\mathbf{W}_{MLP}$ and $\mathbf{W}_{Conv}$ is used to refer to the parameters of the classification layer and the feature extraction layer respectively. Instead of using simple gradient descent,  a recently proposed method for stochastic optimization, the Adam (Adaptive Moment Estimation) algorithm \cite{kingma2014adam}, is utilized for learning the parameters of the network. 

The convolutional layers can be either learned from scratch or finetuned, when a pretrained feature extractor is used. The hidden weights of the MLP are initialized using random orthogonal initialization \cite{saxe2013exact}. The centers of the RBF neurons can be either randomly chosen or initialized using the k-means algorithm over the set of all feature vectors $\mathcal{S} = \{ \mathbf{x}_{ij} | i = 1...N, j = 1 ...N_i \}$. In this work, the set $\mathcal{S}$ is clustered into $N_K$ clusters and the corresponding centroids (codewords) $\mathbf{v}_k\in \mathbb{R}^D (k = 1...N_K)$ are used to initialize the centers of the RBF neurons. The same approach is used in the BoF model to learn the codebook that is used to quantize the feature vectors. However, in contrast to the BoF model, the CBoF model uses this process only to initialize the centers. Both the RBF centers and the scaling factors (initially set to 0.1) are learned.

\subsection {Computational Complexity Analysis}
The asymptotic storage requirements for the network and the cost for the feed-forward process are derived in this Section. Note that these quantities are calculated for the part of the network after the last convolutional layer (assuming that $N_F$ convolutional filters are used in the last convolutional layer,  the size of each feature map/number of feature vectors is $N_i$ and two fully connected layers are used). The proposed method is also compared to a regular CNN as well as to the SPP technique. For both the CBoF and the SPP techniques the number of used spatial regions is denoted by $N_S$. 

The CNN method requires $O(N_iN_FN_H + N_HN_C)$ weights after the last convolutional layer, the SPP method requires $O(N_SN_FN_H + N_HN_C)$ weights, while the proposed CBoF method requires $O(N_SN_KN_F+N_SN_KN_H + N_HN_C)$ weights. Note that the CBoF method is capable of decoupling the input feature dimensionality $N_F$ from the network architecture, allowing to use significantly smaller networks %(usually it is expected that $N_iN_F> N_FN_S > N_KN_S$). 
The cost of feed forwarding the network is $O(N_iN_FN_H + N_HN_C)$ for the CNN, $O(N_iN_F + N_SN_FN_H + N_HN_C)$ for the SPP method and $O(N_iN_KN_F + N_SN_KN_H + N_HN_C)$ for the CBoF method. Both the SPP and the CBoF methods carry an extra cost for the feature aggregation/quantization step ($O(N_iN_F)$ and $O(N_iN_KN_F)$ respectively), but in practice this cost is amortized by the smaller number of operations that are needed after this step (it is reasonable to assume that $N_K < N_H$, as in the conducted experiments). Also, note that since $N_i$ is dependent on the input image size, the cost of the feed forward process can be adjusted for the SPP and the CBoF methods by simply resizing the input image.

\section{Experiments}
\label{section:experiments}

Three different datasets were used to evaluate the proposed method: the MNIST database of handwritten digits (MNIST)~\cite{lecun1998mnist}, the fifteen natural scene (15-scene) dataset~\cite{lazebnik2006beyond}, and the large-scale Annotated Facial Landmarks in the Wild (AFLW) dataset \cite{aflw}.

The MNIST database \cite{lecun1998mnist}, is a well-known dataset that contains 60,000 training images and 10,000 testing images of handwritten digits. The training set was split into 50,000 training images and 10,000 validation images that were used to evaluate the learned models and prevent overfitting. There are 10 different classes, one for each digit (0 to 9), and the size of each image is 28 $\times$ 28.

The 15-scene dataset \cite{lazebnik2006beyond}, consists of images belonging to fifteen different natural scene categories, such as industrial, forest, city, bedroom and living room scenes. The dataset has a total of 4,485 images, with an average size of 300 $\times$ 250 pixels, and the number of images in each category ranges from 200 to 400 images. Since there is no predefined train/test split, the standard evaluation protocol is used \cite{lazebnik2006beyond}:  the train split is composed of 1,500 randomly chosen images (100 from each category), the test split is composed of the remaining 2985 images and the evaluation process is repeated 5 times.

The Annotated Facial Landmarks in the Wild (AFLW) dataset \cite{aflw}, is a large-scale dataset for facial landmark localization. The AFLW dataset was used to evaluate the performance of the proposed method for the problem of facial pose estimation using a light-weight model that can be deployed on embedded devices, such as drones that will assist the video shooting of sport events. Estimating the facial pose of the actors allows for calculating the appropriate shooting angle according to the specifications of each shot. The 75\% of the images were used to train the models, while the rest 25\% for evaluating the accuracy of the models. The face images were cropped according to the given annotation and face images smaller than 16x16 pixels were not used for training or evaluating the model.

In this work, three different feature extraction blocks are utilized to demonstrate the flexibility of the CBoF. For the first one, the \textit{Feature Extraction Layer Block A}, 32 $5\times 5$ convolutional filters are used in the first layer, followed by a $2\times 2$ max layer and 64 $5\times 5$ convolutional filters in the last layer. The \textit{Feature Extraction Layer Block B} is deeper and it is composed of two layers with 64 $3\times 3$  convolutional filters, a layer with 128 3x3 convolutional filters, a $2\times 2$ max-pooling layer followed by another two 256 3x3 convolutional layers, a 512 3x3 convolutional layer and a $2\times 2$ max pooling layer. These blocks are learned from scratch. For the \textit{Feature Extraction Layer Block C} a pretrained convolutional network with 5 convolutional layers is used \cite{zhou2014learning}. This block was pretrained on the Places205 dataset and the feature vectors are extracted from its last convolutional layer (256 feature maps). Different learning rates must be used for each layer, depending on whether the network is learned from scratch or being finetuned. For the first two setups (learning from scratch) the same learning rate is used for all the layers $\eta_{MLP}=\eta_{V}=\eta_{\sigma}=\eta_{conv}=10^{-4}$. For the third setup different learning rates are used, since the feature extraction layer is already trained,: $\eta_{MLP}=\eta_{\sigma}=10^{-4}$, $\eta_{conv}=10^{-5}$ and $\eta_{V}=10^{-2}$.

\subsection {MNIST Evaluation}

\begin{figure}
	\begin{center}
		\includegraphics[width=0.8\linewidth]{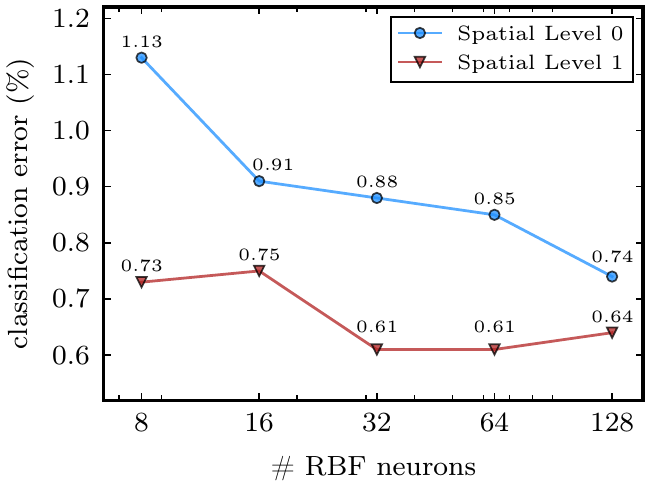}
	\end{center}
	\caption{MNIST: Comparing test error of CBoF for different number of codewords and spatial levels.}
	\label{fig:mnist-spatial-0}
\end{figure}

The well known MNIST dataset is used to study the behavior of the proposed method under different settings. First, the effect of the number of RBF neurons (codewords) is examined in Figure~\ref{fig:mnist-spatial-0} using two different spatial levels, spatial level~0 ($N_S=1$) and spatial level~1  ($N_S=4$) (as shown in Figure~\ref{fig:cbof-spatial}). The network is learned from scratch (feature extraction layer block A). All the layers of the network are trained using back-propagation for 50 epochs. When no spatial segmentation is used (spatial level 0), using more RBF neurons reduces the classification error leading to 0.74\% error when 128 neurons are used. Using spatial segmentation at level 1 further reduces the error to 0.61 when $4\times 32 = 128$ codewords are used. It worths noticing that ``bottlenecking'' the network into just 8 RBF neurons after the convolutional layer, which is less than the classes of the problem,  achieves a remarkable classification error of just 1.13\% (similar to a LeNet-4 \cite{lecun1998gradient}, which fully flattens the feature maps).

Next, the ability of CBoF to learn and classify images of various sizes is examined in Table~\ref{table:mnist-size-training} (32 RBF neurons and spatial level 0 are used). The CBoF is evaluated using images of $20\times 20$, $24\times 24$, $28\times 28$, $32\times 32$ and $36\times 36$ pixels. These images were generated by resizing the original MNIST images (using spline interpolation). Three different training setups are considered: Train A, where only 28x28 images are used for training, Train B, where $24\times 24$, $28\times 28$ and $32\times 32$ images are used for training, and Train C, where all the available image sizes are used for training. When the size of an image to be classified is close to the image size used for training, \ie, $\pm 4 $ pixels or $\pm 14\%$ of training image size, the impact on classification accuracy is relatively small. The scale-invariance of the model is significantly improved when it is trained using images of different sizes (Train B and Train C setups), as shown in Table~\ref{table:mnist-size-training}. Also, the multiple size training process leads to better overall accuracy, since the classification error is reduced to 0.68\% from 0.88\% (for the given setup). 

\begin{figure}
	\begin{center}
		\includegraphics[width=0.32\linewidth]{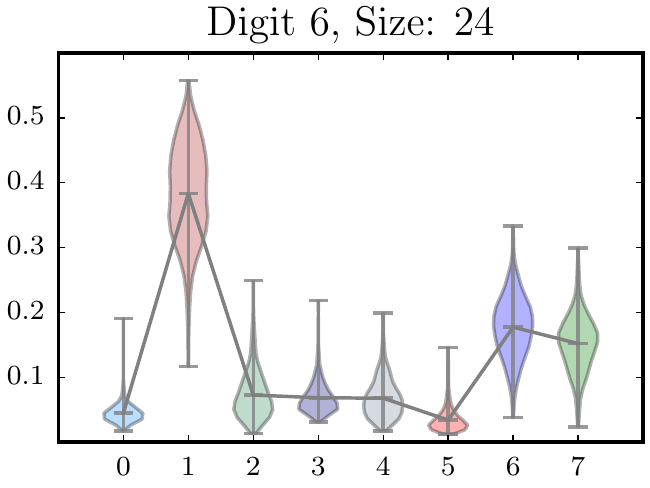}
		\includegraphics[width=0.32\linewidth]{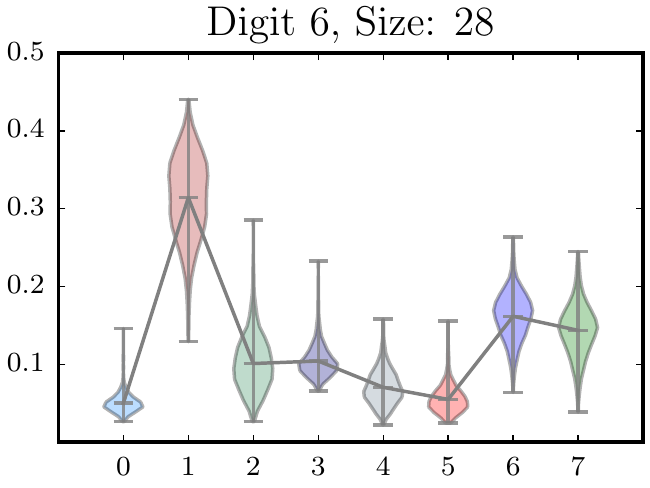}
		\includegraphics[width=0.32\linewidth]{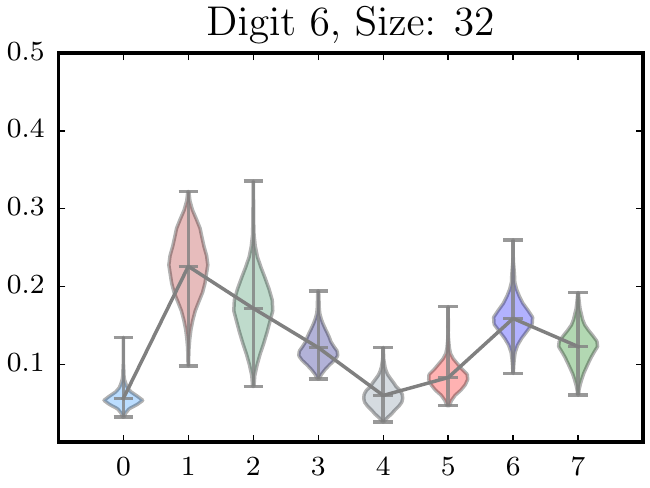}	
	\end{center}
	\caption{MNIST: Visualizing the RBF activations for different image sizes  (without training for different image sizes).}
	\label{fig:mnist-activations-sizes}
\end{figure}

The RBF neurons provide a Voronoi-like segmentation of the space where they operate, \ie, the space of the convolutional feature vectors. This makes the proposed technique relatively invariant to mild distribution shifts. This phenomenon is illustrated in Figure~\ref{fig:mnist-activations-sizes}, where the activations of the RBF neurons are shown when images of different sizes are fed to the network (which is trained only for 28x28 images). Even though the actual input distribution is shifted, relatively mild shifts are observed in the activation of the RBF neurons (especially for the smaller image).

\begin{table}
	\begin{center}
		\begin{tabular}{|l|ccccc|}
			\hline
			\textbf{Image Size}  & 20 & 24 & 28 & 32 & 36 \\
			\hline\hline
			
			Train A & 19.69  &  1.92 &   0.88 &  4.10  & 25.39 \\
			Train B &  3.68 &  0.91 &  0.73  & 1.11  & 2.49 \\
			Train C &  $\mathbf{1.44}$ & $\mathbf{0.82}$ &  $\mathbf{0.68}$  & $\mathbf{0.95}$  & $\mathbf{1.60}$  \\
			\hline
		\end{tabular}
	\end{center}
	\caption{MNIST: CBoF error (\%) for different image sizes and train setups: Train A (28x28 images), Train B ($24\times 24$, $28\times 28$ and $32\times 32$ images) and Train C (all the available image sizes).}
	\label{table:mnist-size-training}
\end{table}

Finally, in Table \ref{table:mnist-final} the CBoF method (spatial level~1 ($N_S=1$), 32 RBF neurons) is compared to other state-of-the art techniques for image classification. For the CNN method the same convolutional layers as in the CBoF method are used, combined with a $2\times 2$ pooling layer (dropout with $p=0.5$ is also used) before the fully connected layer ($1000\times 10$). For the Global Max Polling (GMP)/SPP the global pooling/spatial pooling layer is connected after the last convolutional layer. One spatial level is used for the SPP technique. For the GMP, SPP and the CBoF techniques the models were trained using digit images of $20\times 20$, $24\times 24$, $28\times 28$, $32\times 32$ and $36\times 36$ pixels. The classification error for images of two sizes are reported: for the default image size ($28\times 28$ pixels) and for a reduced image size ($20\times 20$ pixels). The number of parameters used in the layers after the feature extraction layer, which is common among all the evaluated networks, is also reported. All the layers of the networks are learned during the training.

The proposed CBoF outperforms both the CNN and the GMP/SPP techniques, while using significantly less parameters after the convolutional layers (about one order of magnitude less parameters than a similarly performing CNN). Also, it provides much better scale-invariance than the competitive GMP/SPP techniques, even though both techniques were trained using 5 different image sizes. Note that the purpose of the conducted experiments was not to achieve state-of-the-art performance on the MNIST dataset, but to examine the behavior of the proposed method under different training scenarios. Nonetheless, the proposed method achieves performance close to other much more complicated techniques, such as ensembles of neural networks trained using various distortions of the images \cite{ciregan2012multi}. These methods can be combined with the proposed approach and possibly increase the classification accuracy even more, while reducing the overall network size. Also, to demonstrate that the actual architecture of the BoF layer contributes to the improved results, the GMP and SPP networks were also evaluated with an added $1\times 1$ convolutional layer with 64 filters. Again, the proposed CBoF method outperforms both the GMP and SPP methods (3.52\% classification error for the GMP and  1.66\% for the SPP method (for 20x20 images)).

\begin{table}
	\begin{center}
		\begin{tabular}{|l|ccc|}
			\hline
			Method  & Cl.Err. (28) & Cl.Err. (20) & \# Param. \\
			\hline\hline
			CNN & ${0.56}$ & (0.78)* & 1,035k \\
			GMP & 0.62 & 3.22 & 75k \\ 
			SPP & 0.52 & 1.78 & 331k  \\ 
			\hline 
			CBoF(0, 32) & 0.68 & 1.44 & 45k \\
			CBoF(1, 32) & 0.55 & $0.94$ & 147k\\
			CBoF(1, 64) & $\mathbf{0.51}$ & $\mathbf{0.83}$ & 284k\\
			\hline
		\end{tabular}
	\end{center}
	\caption{MNIST: Comparing CBoF to other state-of-the-art techniques. The spatial level and the number of RBF neurons are reported in the parenthesis for the CBoF model. (\footnotesize{*Training from scratch for $20\times 20$ images})}
	
	\label{table:mnist-final}
\end{table}

\subsection {15-scene Evaluation}

\begin{figure}
	\begin{center}
		\includegraphics[width=0.8\linewidth]{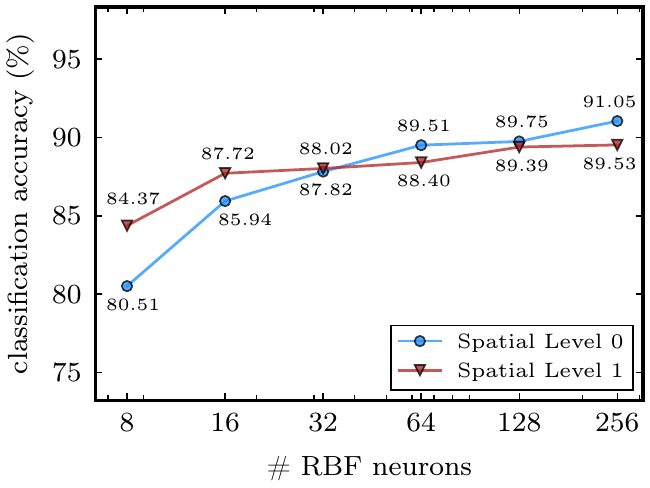}
	\end{center}
	\caption{15-scene: Comparing test accuracy of CBoF for different number of codewords and spatial levels}
	\label{fig:15-scene-spatial-0}
\end{figure}

First, the performance of the CBoF is evaluated for different codebook sizes and spatial levels. Instead of using randomly initialized convolutional layers, a pretrained CNN is used (feature extraction layer block C) \cite{zhou2014learning}. The network was trained for 100 epochs. Also, the fully connected layers were pre-trained for 10 epochs before training the whole network to avoid back-propagating gradients from randomly initialized layers. Figure \ref{fig:15-scene-spatial-0} illustrates the precision on the test set for different number of codewords (evaluated on one test split). As before, using more codewords leads to better classification accuracy when no spatial segmentation is used. Unlike the MNIST dataset, using spatial segmentation does not improve the overall classification accuracy for the 15-scene dataset. This is happening possibly due to the nature of the dataset (the spatial information is less important when recognizing natural scenes than when detecting the edges of aligned digits) and the larger receptive field of the final convolutional filters.

Then, the CBoF model is trained using images of different sizes (one dataset split is used for the evaluation). The results are shown in Table~\ref{table:15scene-size-training}. Again, training with different image sizes helps to improve the classification accuracy for any image size. Note that the classification accuracy increases from 90.66\% to 92.38\%, when multiple image sizes are used for training, highlighting the importance of training the network using different image sizes.

\begin{table}
	\begin{center}
		\begin{tabular}{|l|ccccc|}
			\hline
			\textbf{Im. Size}  & 179 &  203 & 227 & 251 & 275 \\
			\hline\hline
			
			Train A & 88.01 & 90.42 & 90.66 & 90.32 & 88.82 \\
			Train B & 89.66 & 91.61 & 92.01 & 91.45 & 90.03 \\
			Train C & $\mathbf{90.44}$ & $\mathbf{92.05}$ & $\mathbf{92.38}$ & $\mathbf{91.64}$ & $\mathbf{90.39}$  \\
			\hline
		\end{tabular}
	\end{center}
	\caption{15-scene: CBoF accuracy (\%) for different image sizes and train setups: Train A ($227 \times 227$ images), Train B ($203\times 203$, $227\times 227$ and $251 \times  251$ images) and Train C (all the available image sizes).}
	\label{table:15scene-size-training}
\end{table}

Finally, the CBoF (256 RBF neurons, no spatial segmentation) is compared to other state-of-the-art techniques in Table \ref{table:15scene-final}. All the evaluated method share the same feature extraction layer. For the CNN, a $1000\times15$ fully connected layer is added after the last pooling layer, while the GMP/SPP layer is added between the last convolutional and the fully connected layers  Note that the fully connected layer of the pretrained network is not used \cite{zhou2014learning}, leading to slightly different reported results for the CNN method. Instead, all the layers of the network are learned during the training procedure using back-propagation. The following learning rares were used for the CNN, GMP, and SPP methods: $\eta_{conv}=10^{-3}$ and $\eta_{conv}=10^{-5}$. Again, the proposed CBoF method outperforms both the CNN and the GMP/SPP techniques while using significantly less parameters. The same is also true when smaller images are fed to network. The GMP/SPP techniques were also evaluated with one added $1\times 1$ convolutional layer ($86.32\%$ for the GMP, $89.55\%$ for the SPP, 256 $1\times 1$ filters were used). The proposed CBoF method still outperforms there techniques, even though one extra layer was added.

\begin{table}
	\begin{center}
		\begin{tabular}{|p{1.2cm}|ccc|}
			\hline
			Method  & Acc. (227) & Acc. (179) & \# Param. \\
			\hline\hline
			CNN & $ 88.79 \pm 0.62$ & ($87.90 \pm 1.14$*)  & 9,232k\\
			GMP & $ 86.74 \pm 0.40$ & $ 84.62 \pm 0.32$ & 272k \\
			SPP & $ 89.94 \pm 0.73$ & $ 87.70 \pm 0.46$ & 1,296k  \\
			\hline 
			CBoF (0, 256) &$ \mathbf{91.38 \pm 0.63} $ & $ \mathbf{89.86 \pm 0.80}$ & 337k \\
			\hline
		\end{tabular}
	\end{center}
	\caption{15-scene: Comparing CBoF to other state-of-the-art techniques (\footnotesize{*Training from scratch for $179\times 179$ images})} 
		\label{table:15scene-final}
\end{table}

\subsection {AFLW Evaluation}
Finally, the proposed method is evaluated on AFLW dataset. The Feature Extraction Layer Block B was used for these experiments and a hidden layer with $1000\times 1$ neurons was utilized. The models were trained for 5000 iterations, while the CBoF and the GMP/SPP methods were trained using three image sizes, \ie,   $32\times 32$, $64\times 64$, and $96\times 96$ pixels (for each iteration batches from each size were fed to each network). No dropout was used after the feature extraction layer block for the GMP, SPP and CBoF methods, since it was established that slows down the training process significantly. All the networks were trained to perform regression on the horizontal facial pose (yaw) using the mean squared error  as the objective function for the training procedure. The experimental results are shown in Table~\ref{table:aflw-final}. The mean absolute angular error (Ang. Err.) in degrees is reported. As in the previous experiments, the proposed method significantly reduces the size of the models, while still increasing the accuracy of the yaw angle estimation. Note that even though the CBoF (0, 128) and the CBoF (1, 32) use the same number of parameters, the spatial segmentation scheme used in CBoF (1, 32) significantly increases the accuracy since the spatial information is especially important for the task of facial pose estimation.

\begin{table}
	\begin{center}
		\begin{tabular}{|p{1.2cm}|ccc|}
			\hline
			Method  & Ang. Err. (64) & Ang. Err. (32) & \# Param. \\
			\hline\hline
			CNN &  $14.50$ & (13.91*) & 61,954k\\
			GMP & $13.70$ & $12.83$ & 514k \\
			SPP & $12.37$ & $11.96$ & 2,562k \\
			\hline 
			CBoF (0, 128) &  $16.23$ & $13.84$ & 196k \\
			CBoF (1, 32) & $\mathbf{11.92}$  & $\mathbf{11.25}$ & 196k \\	
			\hline
		\end{tabular}
	\end{center}
	\caption{AFLW: Comparing CBoF to other state-of-the-art techniques (the mean absolute angular error in degrees is reported) (\footnotesize{*Training from scratch for $32\times 32$ images})}
	
	\label{table:aflw-final}
\end{table}

\section {Conclusions}
\label{section:conclusions}

In this paper, a neural extension of the well-known BoF model was combined with convolutional neural networks to form powerful image recognition machines.  It was demonstrated, using three image datasets, that the proposed CBoF model is able to natively classify images of various sizes as well as to significantly reduce the number of parameters in the network, while achieving remarkable classification accuracy. In contrast to CNN compression techniques, that simply compress an already trained CNN, the proposed method provides an improved CNN architecture that is end-to-end differentiable and it is invariant to mild distribution shifts. Nonetheless, the proposed technique can be combined with CNN compression techniques, such as \cite{gong2014compressing, han2015deep, wu2015quantized}, to further reduce the size of the resulting network. Furthermore, the proposed RBF-based quantization scheme can be approximated using only linear layers instead of RBF-layers increasing the classification speed  even more. Preliminary experiments show that this approach can indeed increase both the training and the testing speed, without significantly harming the accuracy of the models. The proposed pooling layer can be further enhanced by also learning the spatial regions over which the pooling is performed, similarly to \cite{malinowski2013learnable}, or using visual attention mechanisms \cite{xu2015show}. 

\section*{Acknowledgment}
This project has received funding from the European Union’s Horizon 2020 research and innovation programme under grant agreement No 731667 (MULTIDRONE). This publication reflects the authors' views only. The European Commission is not responsible for any use that may be made of the information it contains.

{\small
	\bibliographystyle{ieee}
	\bibliography{egbib}
}

\end{document}